\documentclass[times]{article}
\author{Christophe Lohou}

\usepackage{arxiv}

\usepackage[utf8]{inputenc} 
\usepackage[T1]{fontenc}    
\usepackage[hidelinks]{hyperref}       
\usepackage{url}            
\usepackage{booktabs}       
\usepackage{amsfonts}       
\usepackage{nicefrac}       
\usepackage{microtype}      
\usepackage{cleveref}       
\usepackage{graphicx}
\usepackage{doi}
\usepackage{longtable}
\usepackage{supertabular}
\usepackage[english]{babel}
\usepackage[table,xcdraw]{xcolor}
\usepackage{breakurl}

\usepackage{amssymb}
\usepackage{latexsym}
\usepackage{psfrag} 
\usepackage{epsfig}

\usepackage{natbib}

\graphicspath{ {./} }
\begin{document}

\title{Topological Classification of points in $\mathcal{Z}^2$\\
by using Topological Numbers\\
for $2$D discrete binary images}
\author{
\href{https://orcid.org/0000-0001-5352-8237}{
    \hspace{1mm}Christophe Lohou}\\
    Université Clermont Auvergne,\\
    Clermont Auvergne INP,\\
    CNRS, Institut Pascal\\
    F-63000 Clermont-Ferrand, France\\
	\texttt{christophe.lohou@uca.fr}
}

\renewcommand{\headeright}{}
\renewcommand{\undertitle}{}
\renewcommand{\shorttitle}{}

\date{\today}
\maketitle

\begin{abstract}

In this paper, we propose a topological classification of points for 2D discrete binary images. This classification is based on the values of the calculus of topological numbers. 
Six classes of points are proposed: isolated point, interior point, simple point, curve point, point of intersection of 3 curves, point of intersection of 4 curves. The number of configurations of each class is also given. 

\end{abstract}


\section{Introduction}

Consider a $2$D binary discrete image consisting of black points (objects) and white points (its complement), points whose coordinates are relative integers. The Digital Topology framework \citep{KoRo89} has made it possible to propose algorithms for simplifying such images by the iterative or parallel deletion of specific points, in order, for example, to produce a smaller form of the objects of the initial image (this is the case with so-called \textit{skeletonization algorithms}). This leads to the notion of simple point, a point is said to be \textit{simple} if its deletion preserves the topology of the image. We have the remarkable property of being able to detect whether a point is simple or not by simply examining a restricted set of points around it (intuitively, its local configuration).  In 3D, a classification of such configurations has been proposed, it is based on topological numbers \citep{MBA1993}.
This classification leads to being able to classify configurations other than those corresponding to simple or non-simple points, for example curve points, surface points \ldots may be detected. 
Such a classification can lead to the retrieval of certain features (either on the initial image  or after a preprocessing step), such as bifurcations on lines in the case of fingerprints images, junctions of arteries in the case of medical angiography images \ldots.

We recently proposed a direct adaptation of topological numbers to the case of 2D discrete binary images \citep{Lohou2024}; in this same study, we also characterized simple points for $2$D binary images with these topological numbers. In this paper, we propose a topological classification of points in a 2D discrete binary image, using these topological numbers. We obtain $6$ classes of points: isolated point, interior point, simple point, curve point, point of intersection of $3$ curves, point of intersection of $4$ curves.
 
The article is written as follows. Basic notions of Digital Topology are recalled in Section \ref{sec:basicNotions}. 
In Sect. \ref{sec:topologyPreservation}, we recall more precisely both the notion of preservation of the topology of an image and the one of a simple point; then we recall both the definition of the topological numbers used for the 2D case, and the characterization of simple points by using topological numbers. 
In Sect. \ref{sec:topologicalClassification}, we propose the classification of the $256$ possible configurations of the neighborhood of any point in an image: $6$ classes are proposed and the numbers of configurations among the $256$ possible ones are given for each class. In Sect. \ref{sec:examples}, we give the  complete classification of two images. 

\section{Basic notions}\label{sec:basicNotions}

\begin{figure}
\psfrag{c}{$a$}
\psfrag{d}{$b$}
\psfrag{e}{$c$}
\psfrag{f}{$d$}
\psfrag{g}{$e$}
\psfrag{h}{$f$}
\psfrag{i}{$g$}
\psfrag{a}{(a)}
\psfrag{b}{(b)}
\begin{center}
\includegraphics[width=8.5cm]{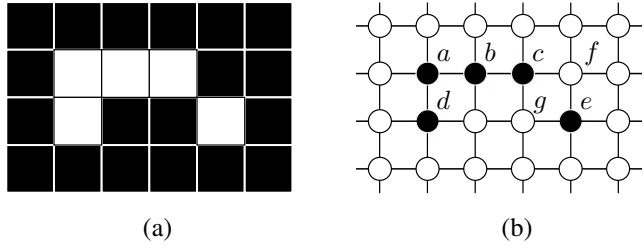}
\end{center}
\caption{(a) A $2$D binary image, (b) a corresponding mapping to $\mathcal{Z}^2$.}
\label{fig:image_2D}
\end{figure}

A $2$D binary image consists of square elements, \textit{pixels},
which can be black for the background and white for the forms of interest, 
also called the \textit{object} afterwards (Fig. \ref{fig:image_2D} (a)). 
A pixel can be mapped to a point in the grid with integer coordinates, 
$\mathcal{Z}^2$, Cartesian product of $\mathcal{Z}$ by $\mathcal{Z}$. 
By convention, we associate in $\mathcal{Z}^2$ 
a black point for a white pixel of the image - thus, to an element of the object -,
and a white point for a black pixel of the image (Fig. \ref{fig:image_2D} (b)).

\begin{figure}
\psfrag{x}{$x$}
\psfrag{a}{(a)}
\psfrag{b}{(b)}
\psfrag{c}{(c)}
\psfrag{d}{(d)}
\begin{center}
\includegraphics[width=9cm]{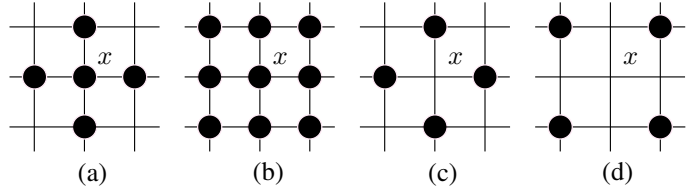}
\end{center}
\caption{(a) $N_4(x)$, (b) $N_8(x)$, (c) $4$-neighbors of $x$, (d) $8$-neighbors of $x$.}
\label{fig:voisinage_2D}
\end{figure}

Let $x(x_1,x_2)$ be the point of coordinates $(x_1,x_2)$ in $\mathcal{Z}^2$. 
Let $x(x_1,x_2)$ and $y (y_1,y_2)$ be two points of $\mathcal{Z}^2$. 
The following two distances can be defined: 
$d_4(x,y)=\sum_{i=1}^{2}|y_i-x_i|$, 
and $d_8(x,y)=\max_{i=1,2} |y_i-x_i|$. 
We define the two following neighborhoods: 
let $x \in \mathcal{Z}^2$, 
$N_4(x)=\{y \in \mathcal{Z}^2, d_4(x,y)\leq 1 \}$ (Fig. \ref{fig:voisinage_2D} (a)), 
and $N_8(x)=\{y \in \mathcal{Z}^2, d_8(x,y)\leq 1 \}$ (Fig. \ref{fig:voisinage_2D} (b)). 
Note $N_n^*(x)=N_n(x) \setminus \{x\}$, for $n \in  \{4,8\}$.  
The $n$ points of $N_n^*(x)$ define the $n$-neighborhood. 
Two points $x$ and $y$ of $\mathcal{Z}^2$ are \textit{$n$-adjacent} if $y \in N_n^*(x)$, $n \in \{4,8\}$. 
The $4$ points of $N_4^*(x)$ are the \textit{$4$-neighbors of $x$} 
(Fig. \ref{fig:voisinage_2D} (c)), 
the $4$ points of $N_8^*(x) \setminus N_4^*(x)$ are the \textit{$8$-neighbors of $x$} 
(Fig. \ref{fig:voisinage_2D} (d)).

Let us recall Jordan's theorem in $\mathcal{R}^2$: 
any simple closed curve separates the plane into two domains which are the interior domain and the exterior domain of the curve.
In order to verify it in the discrete case, we need to use the $4$-adjacency for black points 
and the $8$-adjacency for white points or vice versa \citep{KoRo89}. 
We can then define a \textit{digital} image as the data ($\mathcal{Z}^2, n, \overline{n}, X$),
with $X \subseteq \mathcal{Z}^2$ as the object, 
$(n,\overline{n})=(4,8)$ or $(8,4)$ \citep{KoRo89}; 
the points of $\mathcal{Z}^2$ are the points of the image, 
the points of $X$ are the black points of the image,
and the points of $\overline{X}=\mathcal{Z}^2 \setminus X$, the \textit{complement of $X$} in the image, 
are the white points of the image.

Two black points are \textit{adjacent} if they are $n$-adjacent, 
two white points or one white point and a black point are \textit{adjacent} 
if they are $\overline{n}$-adjacent. A \textit{$n$-path} is a sequence 
$<p_i; 0 \leq i \leq l>$ 
of points such that  
$p_i$ is $n$-adjacent to $p_{i+1}$ for any $0 \leq i < l$; 
if $p_0=p_l$, the path is  said to be \textit{closed}. 
Let $X \subseteq \mathcal{Z}^2$, 
two points $p$ and $q$ of $X$ 
are \textit{$n$-connected into $X$} 
if and only if there is a $n$-path included in $X$ 
which links $p$ to $q$. 
The relation "to be $n$-connected in $X$" is an equivalence relation,
the equivalence classes of this relation are the \textit{$n$-connected components} of the image. 
A $n$-connected component of the set of black points of the image is called a \textit{black component}
and a $\overline{n}$-connected component of the set of white points is called a \textit{white component}.
In a finite digital image (when $X$ is a finite set),
there is a single infinite white component called \textit{background} of the image. 
The finite white components are called \textit{holes}. 

Consider Figure \ref{fig:image_2D} (b). 
If $(n, \overline{n})=(4,8)$, points $c$ and $e$ are not $4$-adjacent, 
there are two $4$-connected components of $X$ ($\{a,b,c,d\},\{e\}$), 
note that points $f$ and $g$ of $\overline{X}$ are $8$-adjacent 
and separate these two $4$-connected components 
(the analog of Jordan's theorem in the $2$D discrete case is verified). 
If $(n,\overline{n})=(8,4)$, 
there is only a single $8$-connected component of $X$ ($\{a,b,c,d,e\}$), 
note that the points $f$ and $g$ of $\overline{X}$ are not $4$-adjacent and 
cannot "cut" the connection between the points $c$ and $e$ of $X$ 
(the analog of Jordan's theorem in the $2$D discrete case is verified). 

\section{Topology preservation, simple points and topological numbers}\label{sec:topologyPreservation}

In this section, we recall the notion of topology preservation and that of simple point. We then recall the definition of topological numbers and the characterization of simple points using topological numbers.

In the remainder of this paper, let $(\mathcal{Z}^2, n, \overline{n}, X)$, with $X \subseteq \mathcal{Z}^2$,  be a digital image.

\subsection{Topology preservation and simple points}

A point $x \in X$ is said to be \textit{$n$-simple for $X$} 
if its deletion from $X$ 
\textit{well preserves the topology of the image}  
in the sense that both 
the number of $n$-connected components of the object $X$ 
and the number of $\overline{n}$-connected components of its complement $\overline{X}$ 
are the same before and after the deletion of the point $x$ \citep{Mor81,KoRo89}. 

\begin{figure}[!t]
\psfrag{x}{$x$}
\psfrag{a}{$a$}
\psfrag{b}{$b$}
\psfrag{c}{$c$}
\psfrag{d}{$d$}
\psfrag{e}{$e$}
\psfrag{f}{$f$}
\psfrag{g}{$g$}
\psfrag{h}{$h$}
\psfrag{i}{(a)}
\psfrag{j}{(b)}
\psfrag{k}{(c)}
\begin{center}
\includegraphics[width=8cm]{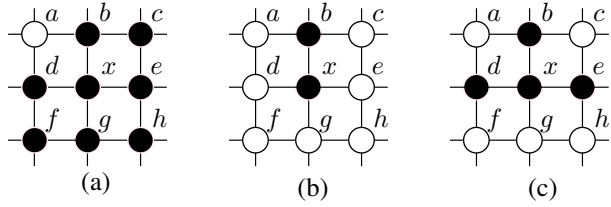}
\end{center}
\caption{(a) $x$ is $4$-simple for $X$ and is not $8$-simple for $X$, 
(b) $x$ is $4$-simple for $X$ and $8$-simple for $X$, 
(c) $x$ is $8$-simple for $X$ and is not $4$-simple for $X$.}
\label{fig:exemples_simples_ou_non_2D}
\end{figure}

Let us consider Fig. \ref{fig:image_2D} (b). 
For $n=4$, 
the image contains two $4$-connected components of $X$ ($\{a,b,c,d\}$ and $\{e\}$),
and a single $8$-connected component of $\overline{X}$. 
After the deletion of the point $a$ from the object $X$, 
the resulting image will contain three $4$-connected components of $X \setminus \{a\}$ 
($\{b,c\}$, $\{d\}$ ,$\{e\}$), thus the topology would not be preserved, therefore the point $a$ is not $4$-simple for $X$.
For $n=8$, 
the image contains a single $8$-connected component of $X$ ($\{a,b,c,d,e\}$),
and a single $4$-connected component of $\overline{X}$. 
After the deletion of $a$ from the object $X$, 
the resulting image will still contain both one $8$-connected component of $X \setminus \{a\}$ 
($\{b,c, d, e\}$), and one $4$-connected component of $\overline{X} \cup \{a\}$. 
Thus the topology would be preserved, therefore the point $a$ is $8$-simple for $X$.

Let us consider another images of Fig. \ref{fig:exemples_simples_ou_non_2D}, that we will also use later:  
\begin{itemize}
\item (a). 
For $n=4$, 
the image contains a single $4$-connected component of $X$ ($\{b,c,d,x,e,f,g,h\}$)
and a single $8$-connected component of $\overline{X}$ ($\{a\}$). 
After the deletion of $x$ from the object $X$, 
the resulting image will contain a single $4$-connected component of $X \setminus \{x\}$ 
($\{b,c,d,e,f,g,h\}$)
and a single $8$-connected component of $\overline{X} \cup \{x\}$ ($\{a,x\}$). 
Thus, both the numbers of $4$-connected components of $X$ and of $X \setminus \{x\}$ 
are unchanged,  
and the numbers of $8$-connected components of $\overline{X}$ 
and of $\overline{X} \cup \{x\}$ are unchanged, 
therefore $x$ is $4$-simple for $X$.   

For $n=8$, 
the image contains a single $8$-connected component of $X$ ($\{b,c,d,x,e,f,g,h\}$)
and a single $4$-connected component of $\overline{X}$ ($\{a\}$). 
After the deletion of $x$ from $X$, 
the resulting image will contain two $4$-connected components of $\overline{X} \cup \{x\}$ 
($\{a\}$ and $\{x\}$, the point $x$ not being $4$-adjacent to the point $a$). 
Since the numbers of white components are not the same before and after the deletion of $x$, 
then the topology can not be preserved,  
therefore the point $x$ is not $8$-simple for $X$. 

\item (b). 
The image contains a single $n$-connected component of $X$ 
and a single $\overline{n}$-connected component of $\overline{X}$, 
the removal of $x$ from $X$ does not change these numbers; 
thus the point $x$ is $n$-simple for $X$, for $n \in \{4,8\}$. 

\item (c). 
The image contains a single $n$-connected component of $X$ ($\{b,d,x,e\}$) 
and three $\overline{n}$-connected components of $\overline{X}$ 
($\{a\}$, $\{c\}$, $\{f, g, h\}$), for $n \in \{4,8\}$.

For $n=4$, the deletion of $x$ from $X$ 
will break the component of $X$ into three $4$-connected components of $X \setminus \{x\}$ 
($\{b\}$, $\{d\}$, $\{e\}$), 
and will merge the three $8$-connected components of $\overline{X}$ 
into a single $8$-connected component of $\overline{X} \cup  \{x\}$ 
($\{a, c, x, f, g, h\}$). 
Thus the topology would not be preserved, 
therefore the point $x$ is not $4$-simple for $X$.  

For $n=8$, after the deletion of $x$ from $X$, 
it remains a single $8$-connected component of $X \setminus \{x\}$ ($\{b,d,e\})$ 
and three $4$-connected components of $\overline{X} \cup \{x\}$ 
($\{a\}$, $\{c\}$, $\{x, f, g, h\}$, 
the point $x$ being $4$-adjacent to the point $g$). 
Thus, both the numbers of $8$-connected components of $X$ and of $X \setminus \{x\}$ 
are the same,  
the numbers of $4$-connected components of $\overline{X}$ and of $\overline{X} \cup \{x\}$ are the same, 
therefore $x$ is $8$-simple for $X$. 
\end{itemize}

\subsection{Topological numbers}

In the previous section, 
in order to check whether a point is simple or not, 
we counted the number of components of the object and its complement in the (overall) image before and after the removal of that point. 
In fact, we have the remarkable property of being able to locally verify 
whether a point $x$ is simple or not 
by the only examination of $N^*_8(x)$ \citep{KoRo89}. 

Bertrand has introduced the topological numbers for $3$D binary images \citep{Be94}: 
topological numbers are the number of connected components of the object and its complement 
in several specific neighborhoods of a point 
and allow us to efficiently check whether a point $x$ is simple or not \citep{BeMa94}. In \citep{Lohou2024}, topological numbers have been defined for $2$D binary images ; 
note that these numbers only use a single neighborhood ($N^*_8(x)$ of a point $x$).
These two numbers also allowed us to characterize simple points in the $2$D case. 
In this section, we recall the $2$D version of these numbers.

Let $X \subset \mathcal{Z}^2$ and $x \in X$. 
Note $C_n(X)$ the number of $n$-connected components of $X$ 
and $C_n^x(X)$ the number of $n$-connected component of $X$ 
and $n$-adjacent to the point $x$, 
the cardinal number of $X$ is denoted by $\#X$. 

\textbf{Definition$1$ (topological numbers in $2$D):} 
Let $X \subset \mathcal{Z}^2$ and $x \in X$. 
The \textit{topological numbers of $X$ and $x$} 
are the two numbers: 
$T_n(x,X)=\#C_n^x[X \cap N^*_8(x)]$, $n \in \{4,8\}$. 

More precisely, 
the topological number for $n=8$ 
is the number $T_8(x,X)=\#C_8^x[X \cap N^*_8(x)]=\#C_8[X \cap N^*_8(x)]$ 
(since a $8$-connected component included in $N^*_8(x)$ is necessarily $8$-adjacent to $x$); 
the topological number for $n=4$ 
is the number $T_4(x,X)=\#C_4^x[X \cap N^*_8(x)]$. 

\begin{table}
\centering
\renewcommand{\arraystretch}{1.5}
\begin{tabular}{|l||c|c|c|c|c|c|}
\hline
      & $k=0$ & $k=1$ & $k=2$ & $k=3$ & $k=4$ & $k>4$\\ 
\hline
$T_4(x,X)=k$ & $16$ & $117$ & $102$ & $20$ & $1$  & $0$\\
\hline
$T_8(x,X)=k$ & $1$ & $132$ & $102$ & $20$  & $1$ & $0$\\
\hline
\hline
      & $k=0$ & $k=1$ & $k=2$ & $k=3$ & $k=4$ & $k>4$\\ 
\hline
$T_8(x,\overline{X})=k$ & $1$ & $132$ & $102$ & $20$  & $1$ & $0$\\
\hline
$T_4(x,\overline{X})=k$ & $16$ & $117$ & $102$ & $20$ & $1$  & $0$\\
\hline
\end{tabular}
\vspace*{0.25cm}
\caption{Number of configurations in $N^*_8(x)$ of a point $x$ which belongs to $X$, 
such that $T_n(x,X)=k$ or $T_{\overline{n}}(x,\overline{X})=k$ with $k \in \mathbb{N}$, for $n \in \{4, 8\}$. 
}
\label{tab:tab_nombre_topo_T_ou_Tbarre}
\end{table}

In the following, we call \textit{(local) configuration} of a point $x$, the neighborhood $N^*_8(x)$ of a point $x$.

For the $2$D case, by reviewing with a computer 
all $2^8=256$ configurations of black and white points 
in the neighborhood $N^*_8(x)$ of a point $x$, 
we have computed the number of configurations having specific values of $T_n(x,X)$ or $T_{\overline{n}}(x,\overline{X})$
for $n \in \{4, 8\}$, see Tab. \ref{tab:tab_nombre_topo_T_ou_Tbarre} \citep{Lohou2024}.

\subsection{Local characterization of simple points}\label{sec:localcharacterization}

In $3$D, we have the property that a point is simple if and only if its two topological numbers are both equal to one.

By reviewing with a computer 
all $256$ configurations of black and white points 
in the neighborhood $N^*_8(x)$ of a point $x$, 
we have verified that any configuration such that the topology is well preserved 
before and after the deletion of $x$ 
(by verifying the preservation of numbers of connected components 
of an object $X$ and its complement $\overline{X}$ 
before and after 
the deletion of the point $x$) 
if and only if both $T_n(x,X)=1$ and $T_{\overline{n}}(x,\overline{X})=1$, 
for $n \in \{4, 8\}$ \citep{Lohou2024}. 
We highlight that for the $2$D case, only one neighborhood ($N^*_8(x)$ of a point $x$) is sufficient to check whether a point is simple or not, in contrary to the $3$D case.

Therefore, we have the following Proposition:

\textbf{Proposition$1$ (local characterization of a $n$-simple point):}  
Let $X \subseteq \mathcal{Z}^2$ and $x \in X$. 
The point $x$ is a $n$-simple point for $X$ $\Leftrightarrow$ $T_n(x,X)=1$ and $T_{\overline n}(x,\overline{X})=1$. 

Let us consider again the examples of Fig. \ref{fig:exemples_simples_ou_non_2D} 
to check the simplicity or not of points $x$ 
using topological numbers:
\begin{itemize}
\item (a). 
$N^*_8(x)\cap X$ contains a single $4$-connected component (thus $8$-connected too) of $X$ 
($\{b,c,d,e,f,g,h\}$), 
this component is $4$-adjacent to $x$, 
thus $T_4(x,X)=T_8(x,X)=1$.
$N^*_8(x) \cap \overline{X}$ contains only one $4$-connected component (thus $8$-connected too) 
of $\overline{X}$ ($\{a\})$, 
this component is not $4$-adjacent to $x$, 
however this component is $8$-adjacent to $x$, 
thus $T_4(x,\overline{X})=0$ and $T_8(x,\overline{X})=1$. 
Therefore, the point $x$ is $4$-simple for $X$, but is not $8$-simple for $X$. 

\item (b). 
$N^*_8(x) \cap X$ contains a single $4$-connected component 
(thus $8$-connected too) of $X$ ($\{b\}$)  
and this component is $4$-adjacent to the point $x$, thus $T_4(x,X)=T_8(x,X)=1$. 
$N^*_8(x) \cap \overline{X}$ contains a single $4$-connected component (thus $8$-connected too) 
of $\overline{X}$ ($\{a, c, d, e, f, g, h\}$) and it is $4$-adjacent to the point $x$;  
therefore $T_4(x,\overline{X})=T_8(x,\overline{X})=1$. 
The point $x$ is $4$-simple for $X$ and is $8$-simple for $X$ too. 

\item (c). 
For $n=4$, $N^*_8(x) \cap X$ is made of 
three $4$-connected components of $X$ that are $4$-adjacent to $x$ ($\{b\}, \{d\}, \{e\}$)
and $N^*_8(x) \cap \overline{X}$ contains three $8$-connected components of $\overline{X}$ 
that are $8$-adjacent to $x$ 
($\{a\}, \{c\}, \{f,g,h\}$)
thus $T_4(x,X)=3$ and $T_8(x,\overline{X})=3$, 
therefore the point $x$ is not $4$-simple for $X$. 
\\
For $n=8$, $N^*_8(x) \cap X$ is made of 
a single $8$-connected component of $X$ ($\{b,d,e\}$) 
and $N^*_8(x) \cap \overline{X}$ is made of three $4$-connected components of $\overline{X}$ 
($\{a\}, \{c\}, \{f,g,h\}$), 
but only the component $\{f,g,h\}$ is $4$-adjacent to $x$; 
thus $T_8(x,X)=1$ and $T_4(x,\overline{X})=1$, 
therefore the point $x$ is $8$-simple for $X$. 
\end{itemize}

\section{Topological classification of points of $\mathcal{Z}^2$}\label{sec:topologicalClassification}

In \citep{MBA1993}, a topological classification of points for $3$D binary images 
has been proposed by distinguishing values of pairs $(T_n(x,X), T_{\overline{n}}(x,\overline{X}))$.
Here, by reviewing with a computer 
all configurations of black and white points 
in the neighborhood $N^*_8(x)$ of a point $x$, 
we may also propose a classification of points for $2$D binary images 
by using topological numbers given in Definition1. 

\begin{table}
\centering
\renewcommand{\arraystretch}{1.5}
\begin{tabular}{|l||c|c|c|c|c|}
\hline
$(n,\overline{n})=(4,8)$ & $T_{\overline{n}}(x,\overline{X})=0$ & $T_{\overline{n}}(x,\overline{X})=1$ & $T_{\overline{n}}(x,\overline{X})=2$ & $T_{\overline{n}}(x,\overline{X})=3$ & $T_{\overline{n}}(x,\overline{X})=4$ \\ 
\hline
$T_n(x,X)=0$ & $0$ & $16$ & $0$ & $0$ & $0$ \\
\hline
$T_n(x,X)=1$ & $1$ & $116$ & $0$ & $0$ & $0$ \\
\hline
$T_n(x,X)=2$ & $0$ & $0$ & $102$ & $0$ & $0$ \\
\hline
$T_n(x,X)=3$ & $0$ & $0$ & $0$ & $20$ & $0$ \\
\hline
$T_n(x,X)=4$ & $0$ & $0$ & $0$ & $0$ & $1$ \\
\hline
\hline
$(n,\overline{n})=(8,4)$ & $T_{\overline{n}}(x,\overline{X})=0$ & $T_{\overline{n}}(x,\overline{X})=1$ & $T_{\overline{n}}(x,\overline{X})=2$ & $T_{\overline{n}}(x,\overline{X})=3$ & $T_{\overline{n}}(x,\overline{X})=4$ \\ 
\hline
$T_n(x,X)=0$ & $0$ & $1$ & $0$ & $0$ & $0$ \\
\hline
$T_n(x,X)=1$ & $16$ & $116$ & $0$ & $0$ & $0$ \\
\hline
$T_n(x,X)=2$ & $0$ & $0$ & $102$ & $0$ & $0$ \\
\hline
$T_n(x,X)=3$ & $0$ & $0$ & $0$ & $20$ & $0$ \\
\hline
$T_n(x,X)=4$ & $0$ & $0$ & $0$ & $0$ & $1$ \\
\hline
\end{tabular}
\vspace*{0.25cm}
\caption{Number of configurations (in $N^*_8(x)$) of a point $x$ which belongs to $X$, 
such that $T_n(x,X)=k$ and $T_{\overline{n}}(x,\overline{X})=k'$ 
with $(k,k') \in \mathbb{N}^2$, for $n \in \{4, 8\}$. 
}
\label{tab:tab_nombre_topo_T_et_Tbarre}
\end{table}

\begin{figure*}[!h]
\psfrag{T11}{only for $(n,\overline{n})=(8,4)$}
\psfrag{T10}{for $(n,\overline{n})=(4,8)$ and $8,4)$}
\psfrag{T9}{only for $(n,\overline{n})=(4,8)$}
\psfrag{T1}{$\#(T_n(x,X)=0, T_{\overline{n}}(x,\overline{X})=1)$}
\psfrag{T2}{$\#(T_n(x,X)=1, T_{\overline{n}}(x,\overline{X})=0)$}
\psfrag{T3}{$\#(T_n(x,X)=1, T_{\overline{n}}(x,\overline{X})=1)$}
\psfrag{T4}{$\#(T_n(x,X)=2, T_{\overline{n}}(x,\overline{X})=2)$}
\psfrag{T5}{$\#(T_n(x,X)=3, T_{\overline{n}}(x,\overline{X})=3)$}
\psfrag{T6}{$\#(T_n(x,X)=4, T_{\overline{n}}(x,\overline{X})=4)$}
\psfrag{T7}{$=16$ for $(n,\overline{n})=(4,8)$}
\psfrag{T8}{$=1$ for $(n,\overline{n})=(8,4)$}
\psfrag{T9}{$=1$ for $(n,\overline{n})=(4,8)$}
\psfrag{T10}{$=16$ for $(n,\overline{n})=(8,4)$}
\psfrag{T11}{$=116$}
\psfrag{T12}{$=102$}
\psfrag{T13}{$=20$}
\psfrag{T14}{$=1$}
\psfrag{conn1}{$(n,\overline{n})=(4,8)$}
\psfrag{conn2}{$(n,\overline{n})=(8,4)$}
\psfrag{isolated}{$n$-isolated point}
\psfrag{interior}{$n$-interior point}
\psfrag{simple}{$n$-simple point}
\psfrag{curve}{$n$-curve point}
\psfrag{curv3}{$(n,3)$-curves junction point}
\psfrag{curv4}{$(n,4)$-curves junction point}
\psfrag{y}{$\emptyset$}
\psfrag{x}{$x$}
\psfrag{onlyfor}{only for}
\psfrag{for}{for}
\psfrag{and}{and}
\psfrag{a}{(a)}
\psfrag{b}{(b)}
\psfrag{c}{(c)}
\psfrag{d}{(d)}
\psfrag{e}{(e)}
\psfrag{f}{(f)}
\psfrag{g}{(g)}
\psfrag{h}{(h)}
\psfrag{i}{(i)}
\psfrag{j}{(j)}
\psfrag{k}{(k)}
\psfrag{l}{(l)}
\psfrag{m}{(m)}
\psfrag{n}{(n)}
\psfrag{o}{(o)}
\psfrag{p}{(p)}
\psfrag{q}{(q)}
\psfrag{r}{(r)}
\begin{center}
\includegraphics[width=14.5cm]{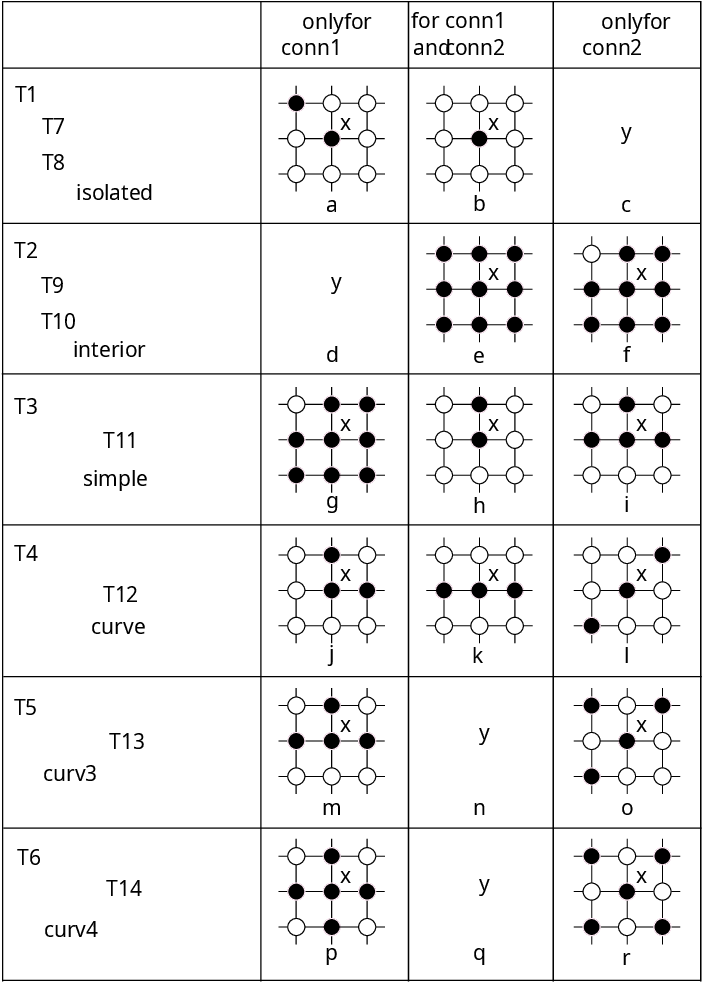}
\end{center}
\caption{Number of configurations for the different values of topological numbers 
and according to the values  $(n,\overline{n})$. 
In the last three columns, 
some examples of configurations are given  
for each possibility of values for the pair ($T_n(x,X), T_{\overline{n}}(x,\overline{X})$) 
by specifying 
if there are any configuration verifying these pairs only for $n=4$, both for $n=4$ and $n=8$, 
or only for $n=8$.}
\label{fig:table_nombres_topos_2D}
\end{figure*}

Among the $2^8=256$ possible configurations of black and white points 
that belong to the neighborhood $N^*_8(x)$ of a point $x$, 
there are only $6$ possible values of pairs of topological numbers 
$(T_n(x,X), T_{\overline{n}}(x,\overline{X}))$, see Tab. \ref{tab:tab_nombre_topo_T_et_Tbarre}. 
The number of configurations for each class defined by the values of the pair $(T_n(x,X),T_{\overline{n}}(x,\overline{X}))$; 
some examples of configurations for each class 
are given in Fig. \ref{fig:table_nombres_topos_2D}. 

\newpage
Therefore, we may propose the following definition of the topological classification of points for $2$D discrete binary images:

\textbf{Definition$2$ (topological classification for $2$D discrete binary images):} 
Let $X \subset \mathcal{Z}^2$ and $x \in X$, the point $x$ belongs to exactly a single class amongst the $6$ following ones 
according to its topological numbers $T_n(x,X)$ and $T_{\overline{n}}(x,\overline{X})$:
\begin{itemize} 
\item $T_n(x,X)=0$ and $T_{\overline{n}}(x,\overline{X})=1$ 
$\Leftrightarrow$ no point of $X$ is $n$-adjacent to $x$ in $N^*_8(x)$
and there is a single $\overline{n}$-connected component of $\overline{X}$ 
and $\overline{n}$-adjacent to $x$
$\Leftrightarrow$ $x$ is said to be a \textit{$n$-isolated point} of $X$. 
There are $16$ configurations that correspond to a $4$-isolated point 
(one of them is given in Fig. \ref{fig:table_nombres_topos_2D} (a)); 
only a single configuration corresponds to a $8$-isolated point 
(Fig. \ref{fig:table_nombres_topos_2D} (b)),  
\item $T_n(x,X)=1$ and $T_{\overline{n}}(x,\overline{X})=0$ 
$\Leftrightarrow$ no point of $\overline{X}$ is $\overline{n}$-adjacent to $x$ in $N^*_8(x)$ 
and there is a single $n$-connected component of $X$ 
and $n$-adjacent to $x$ 
$\Leftrightarrow$ $x$ is said to be a \textit{$n$-interior point} of $X$. 
There are $16$ configurations that correspond to a $8$-interior point 
(one of them is given in Fig. \ref{fig:table_nombres_topos_2D} (f)); 
only a single configuration corresponds to a $4$-interior point 
(Fig. \ref{fig:table_nombres_topos_2D} (e)),  
\item $T_n(x,X)=1$ and $T_{\overline{n}}(x,\overline{X})=1$, examples are given in 
Fig. \ref{fig:table_nombres_topos_2D} (g) to (i); in Sec. \ref{sec:localcharacterization}, 
we have shown that such configurations correspond to a 
\textit{$n$-simple point} for $X$. 
There are $116$ configurations corresponding to a $n$-simple point for $X$, for $n \in \{4, 8\}$,
\item $T_n(x,X)=2$ and $T_{\overline{n}}(x,\overline{X})=2$, examples are given in 
Fig. \ref{fig:table_nombres_topos_2D} (j) to (l),
$x$ is said to be a \textit{$n$-curve point} of $X$.  
There are $102$ configurations which correspond to a $n$-curve-end point of $X$, 
for $n \in \{4, 8\}$,
\item $T_n(x,X)=3$ and $T_{\overline{n}}(x,\overline{X})=3$, 
$x$ is said to be a \textit{$(n,3)$-curves junction point} of $X$, 
examples are given in Fig. \ref{fig:table_nombres_topos_2D} (m) and (o). 
There are $20$ configurations which correspond to a $(n,3)$-curves junction point, for $n \in \{4, 8\}$,
\item $T_n(x,X)=4$ and $T_{\overline{n}}(x,\overline{X})=4$, 
$x$ is said to be a \textit{$(n,4)$-curves junction point} of $X$, 
examples are given in Fig. \ref{fig:table_nombres_topos_2D} (p) and (r). 
There is a single configuration which corresponds to a $(n,4)$-curves junction point, for $n \in \{4, 8\}$.

\end{itemize}

Let us consider again the examples of Fig. \ref{fig:exemples_simples_ou_non_2D} :
\begin{itemize}
\item (a). 
$T_4(x,X)=T_8(x,X)=1$, 
$T_4(x,\overline{X})=0$ and $T_8(x,\overline{X})=1$. 
Therefore, the point $x$ is $4$-simple for $X$ 
(this is also the configuration depicted in Fig. \ref{fig:table_nombres_topos_2D} (g), one amongst the $116$ possible  configurations in $N^*_8(x)$ which correspond to a $4$-simple point).
The point $x$ is not $8$-simple for $X$ 
(this is also the configuration depicted in Fig. \ref{fig:table_nombres_topos_2D} (f), one amongst the $16$ possible  configurations in $N^*_8(x)$ which correspond to a $8$-interior point).
  
\item (b). 
$T_4(x,X)=T_8(x,X)=1$, 
$T_4(x,\overline{X})=T_8(x,\overline{X})=1$. 
The point $x$ is $4$-simple for $X$ and is $8$-simple for $X$ too. 
The configuration of $x$ is also the one depicted in Fig. \ref{fig:table_nombres_topos_2D} (h), one amongst the $116$ possible configurations in $N^*_8(x)$ which correspond to either a $4$-simple point, or a $8$-simple point or a both $4$-simple and $8$-simple point - which is the case here.

\item (c). 
For $n=4$, $T_4(x,X)=3$ and $T_8(x,\overline{X})=3$, 
therefore the point $x$ is not $4$-simple for $X$. 
This is also the configuration of Fig. \ref{fig:table_nombres_topos_2D} (m), one amongst the $20$ possible  configurations in $N^*_8(x)$ which correspond to a $(4,3)$-curves junction point.
\\
For $n=8$, 
$T_8(x,X)=1$ and $T_4(x,\overline{X})=1$, 
therefore the point $x$ is $8$-simple for $X$. 
This is also the configuration depicted in Fig. \ref{fig:table_nombres_topos_2D} (i). 

\end{itemize}

\clearpage

\section{Examples}\label{sec:examples}

\begin{figure}[!t]
\psfrag{x1}{$x_1$}
\psfrag{x2}{$x_2$}
\psfrag{x3}{$x_3$}
\psfrag{x4}{$x_4$}
\psfrag{x5}{$x_5$}
\psfrag{x6}{$x_6$}
\begin{center}
\includegraphics[width=6.5cm]{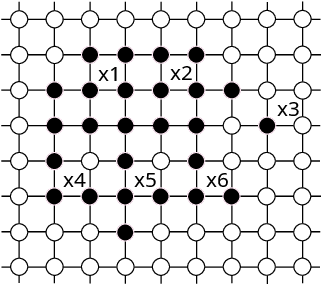}
\end{center}
\caption{
The point $x_1$ is a $4$-simple point (Fig. \ref{fig:table_nombres_topos_2D} (g)), 
$x_2$ is the $4$-interior point (Fig. \ref{fig:table_nombres_topos_2D} (e)), 
$x_3$ is a $4$-isolated point (Fig. \ref{fig:table_nombres_topos_2D} (a)), 
$x_4$ is a $4$-curve point (Fig. \ref{fig:table_nombres_topos_2D} (j)), 
$x_5$ is the $(4,4)$-curves junction point (Fig. \ref{fig:table_nombres_topos_2D} (p)), 
$x_6$ is a $(4,3)$-curves junction point (Fig. \ref{fig:table_nombres_topos_2D} (m)).}
\label{fig:classification1_conn4}
\end{figure}

\begin{figure}[!t]
\psfrag{x1}{$x_1$}
\psfrag{x2}{$x_2$}
\psfrag{x3}{$x_3$}
\psfrag{x4}{$x_4$}
\psfrag{x5}{$x_5$}
\psfrag{x6}{$x_6$}
\begin{center}
\includegraphics[width=7.5cm]{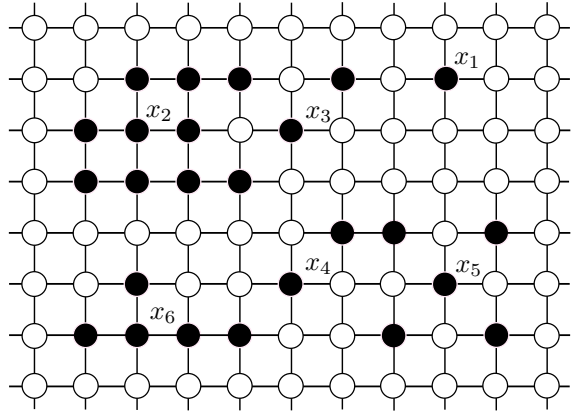}
\end{center}
\caption{
The point $x_1$ is the $8$-isolated point (Fig. \ref{fig:table_nombres_topos_2D} (b)), 
$x_2$ is a $8$-interior point (Fig. \ref{fig:table_nombres_topos_2D} (f)), 
$x_3$ is a $(8,3)$-curves junction point (Fig. \ref{fig:table_nombres_topos_2D} (o)), 
$x_4$ is a $8$-curve point (Fig. \ref{fig:table_nombres_topos_2D} (l)), 
$x_5$ is the $(8,4)$-curves junction point (Fig. \ref{fig:table_nombres_topos_2D} (r)), 
$x_6$ is a $8$-simple point (Fig. \ref{fig:table_nombres_topos_2D} (i)).
}
\label{fig:classification1_conn8}
\end{figure}

In Fig. \ref{fig:classification1_conn4} (resp. Fig. \ref{fig:classification1_conn8}), 
we give an image such that each of any possible topological classes for points of $X$ defined with these two topological numbers occurs:
$n$-isolated point, $n$-interior point, $n$-simple point, 
$n$-curve point, $(n,3)$-curves junction point, $(n,4)$-curves junction point
for $n=4$ (resp. $n=8$), by using several configurations given in Tab. \ref{fig:table_nombres_topos_2D}.

\begin{figure}
\begin{center}
\psfrag{4-isole}{$4$-isolated point}
\psfrag{4-interieur}{$4$-interior point}
\psfrag{4-simple}{$4$-simple point}
\psfrag{4-courbe}{$4$-curve point}
\psfrag{4-3-curve}{$(4,3)$-curves}
\psfrag{4-4-curve}{$(4,4)$-curves}
\psfrag{junction-point}{junction point}
\includegraphics[width=12cm]{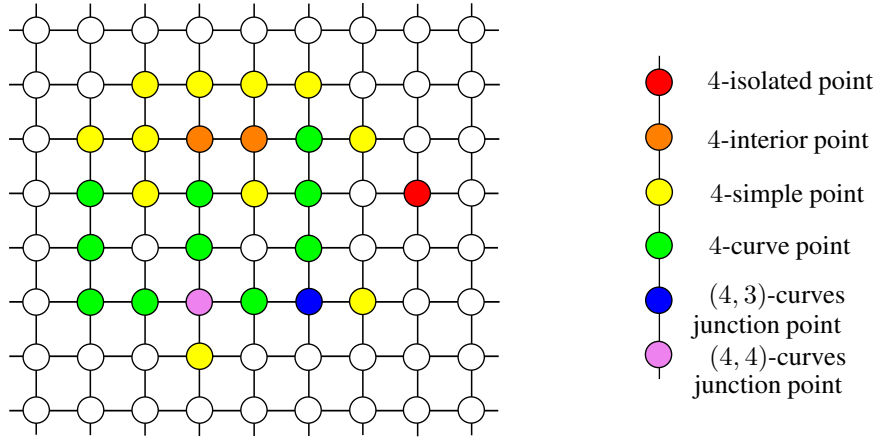}
\end{center}
\caption{
Topological classification of the image of Fig. \ref{fig:classification1_conn4}, 
by using the color code for any possible topological class, for $n=4$, depicted in the colormap 
in the right of the figure.}
\label{fig:classification2_conn4}
\end{figure}

\begin{figure}
\begin{center}
\psfrag{8-isole}{$8$-isolated point}
\psfrag{8-interieur}{$8$-interior point}
\psfrag{8-simple}{$8$-simple point}
\psfrag{8-courbe}{$8$-curve point}
\psfrag{8-3-curve}{$(8,3)$-curves}
\psfrag{8-4-curve}{$(8,4)$-curves}
\psfrag{junction-point}{junction point}
\includegraphics[width=14cm]{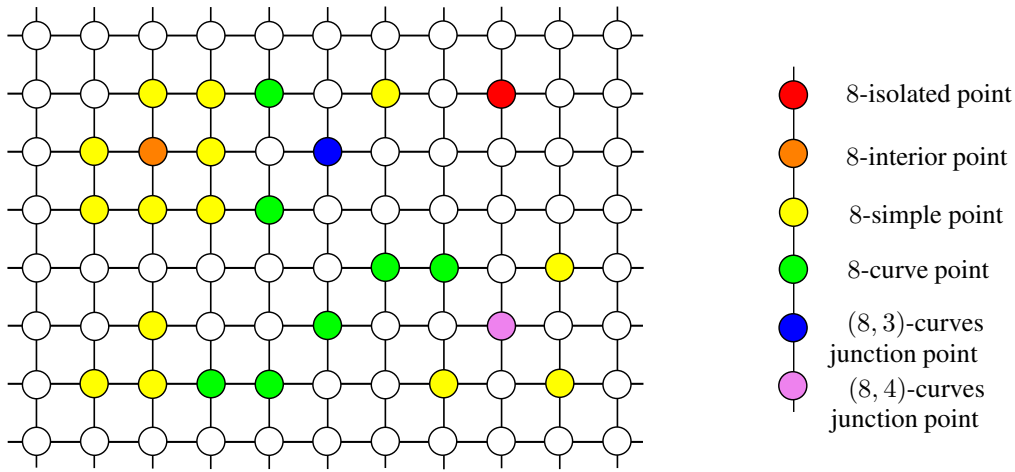}
\end{center}
\caption{Topological classification of the image of Fig. \ref{fig:classification1_conn8}, 
by using the color code for any possible topological class, for $n=8$, depicted in the colormap 
in the right of the figure.
}
\label{fig:classification2_conn8}
\end{figure}

In Fig. \ref{fig:classification2_conn4} (resp. Fig. \ref{fig:classification2_conn8}), we propose the topological
classification of all points of the image of Fig. \ref{fig:classification1_conn4} 
(resp. Fig. \ref{fig:classification1_conn8})
by using a color code different for each of the $6$ topological classes.

\section{Conclusion}

The direct use of topological numbers allows us to propose a classification in $6$ classes: isolated point, interior point, simple point, curve point, point of intersection of 3 curves, point of intersection of 4 curves. 

Note that we can divide a topological class in several subclasses: 
for example, we may distinguish curve-end points (they correspond to simple points with only one neighbor in $X$ in $N^*_8(x)$, 
and $n$-adjacent to $x$, for any point $x$) and non curve-end points, from the topological class that includes simple points. 
Thus, during a deleting process of simple points, if curve-end points are prevented to be deleted, then present curves or occuring curves are kept to yield a curve skeleton. 
Finally, such a classification can be useful when extracting features as in the case of fingerprint images, medical images.

\bibliographystyle{unsrtnat}

\end{document}